\documentclass[conference]{IEEEtran}
\IEEEoverridecommandlockouts

\usepackage{cite}
\usepackage{amsmath,amssymb,amsfonts}
\usepackage{algorithmic}
\usepackage{graphicx}
\usepackage{textcomp}
\usepackage{xcolor}
\usepackage{afterpage}
\usepackage{tikz}
\usepackage{subcaption}
\usepackage{listings}
\usepackage{hyperref}

\definecolor{keywordcolor}{RGB}{0,119,170}  
\definecolor{stringcolor}{RGB}{186,33,33}   
\definecolor{commentcolor}{RGB}{24,128,24}  
\definecolor{functioncolor}{RGB}{169,13,145}  
\def\BibTeX{{\rm B\kern-.05em{\sc i\kern-.025em b}\kern-.08em
    T\kern-.1667em\lower.7ex\hbox{E}\kern-.125emX}}
\begin{document}

\title{DaCe AD: Unifying High-Performance Automatic Differentiation for Machine Learning and Scientific Computing\\
}

\author{\IEEEauthorblockN{Afif Boudaoud}
\IEEEauthorblockA{
\textit{ETH Zurich}\\
afif.boudaoud@inf.ethz.ch}
\and
\IEEEauthorblockN{Alexandru Calotoiu}
\IEEEauthorblockA{
ETH Zurich \\
alexandru.calotoiu@inf.ethz.ch}
\and
\IEEEauthorblockN{Marcin Copik}
\IEEEauthorblockA{
ETH Zurich \\
marcin.copik@inf.ethz.ch}
\and
\IEEEauthorblockN{Torsten Hoefler}
\IEEEauthorblockA{
ETH Zurich \\
torsten.hoefler@inf.ethz.ch}
}

\maketitle

\begin{abstract}
Automatic differentiation (AD) is a set of techniques that systematically applies the chain rule to compute the gradients of functions without requiring human intervention. Although the fundamentals of this technology were established decades ago, it is experiencing a renaissance as it plays a key role in efficiently computing gradients for backpropagation in machine learning algorithms. AD is also crucial for many applications in scientific computing domains, particularly emerging techniques that integrate machine learning models within scientific simulations and schemes. Existing AD frameworks have four main limitations: limited support of programming languages, requiring code modifications for AD compatibility, limited performance on scientific computing codes, and a naive store-all solution for forward-pass data required for gradient calculations. These limitations force domain scientists to manually compute the gradients for large problems. This work presents DaCe AD, a general, efficient automatic differentiation engine that requires no code modifications. DaCe AD uses a novel ILP-based algorithm to optimize the trade-off between storing and recomputing to achieve maximum performance within a given memory constraint. We showcase the generality of our method by applying it to NPBench, a suite of HPC benchmarks with diverse scientific computing patterns, where we outperform JAX, a Python framework with state-of-the-art general AD capabilities, by more than 92 times on average without requiring any code changes.
\end{abstract}


\section{Introduction}
Automatic differentiation (AD) is a technique for calculating the derivatives of programs~\cite{art_ad} that outperforms traditional methods like symbolic and numerical differentiation, particularly for complex algorithms and mathematical functions~\cite{ad_review}. AD is critical for training neural networks, as it computes gradients required for backpropagation~\cite{Rumelhart1986LearningRB,30_years_of_adaptive_nn}. Advancements in deep learning were significantly facilitated by AD's efficiency in computing gradients for complex loss functions~\cite{BaydinPR15}, enabling the training of large-scale models such as modern Large Language Models~\cite{vaswani2023attentionneed}. Beyond the confines of machine learning in fields such as atmospheric sciences and oceanography, AD is crucial for sensitivity analysis~\cite{christopher2002identification}, parameter estimation~\cite{beck1977parameter}, and data assimilation~\cite{bouttier2002data}. Furthermore, many recent scientific algorithms have integrated machine learning models within their systems~\cite{kochkov2024neuralgeneralcirculationmodels}, making AD more relevant even for traditional applications such as weather simulation~\cite{alley2019advances}.

\begin{figure}
    \centering
    \includegraphics[width=1\linewidth]{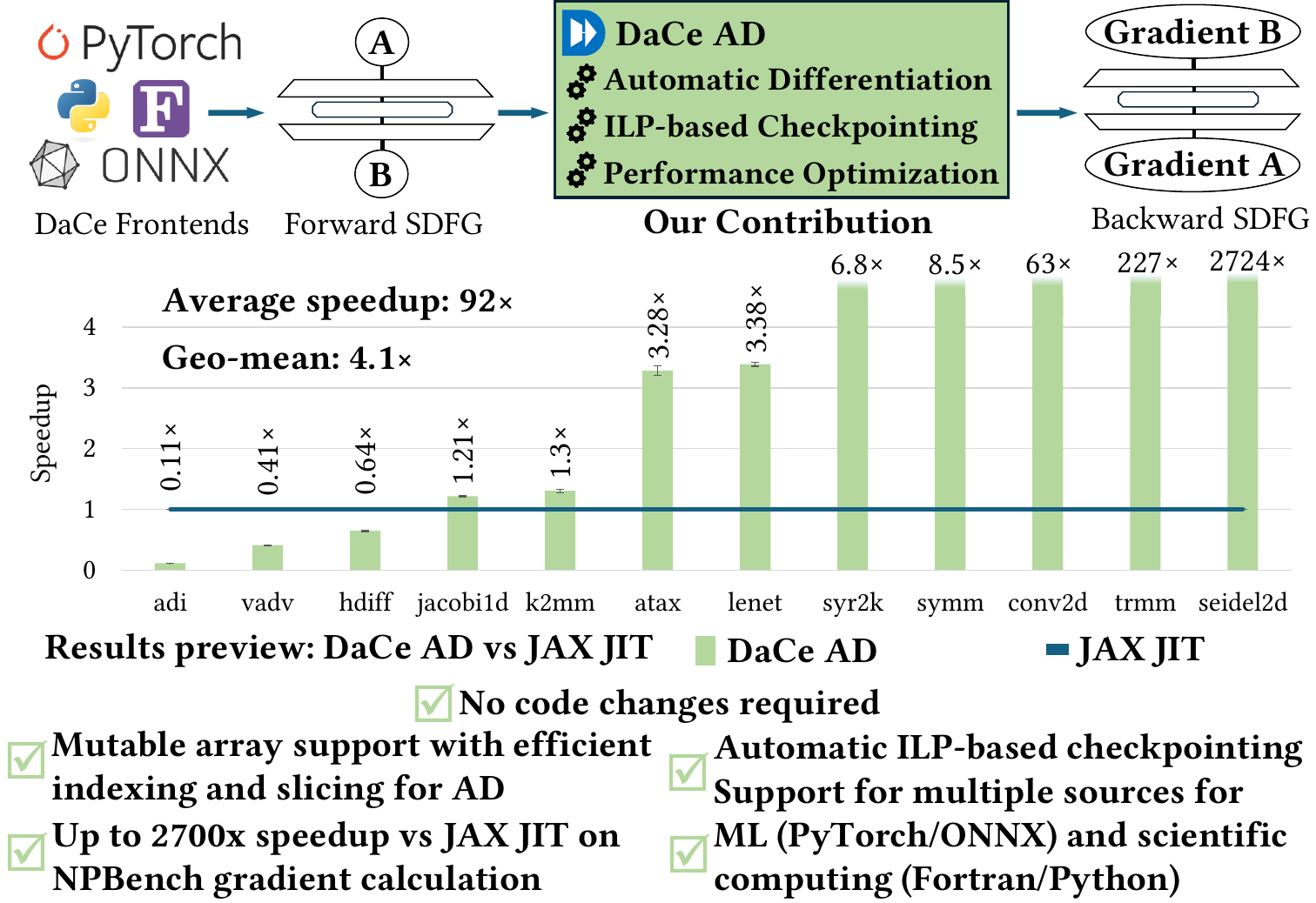}
    \caption{DaCe AD Contribution Overview.}
    \label{fig:dacead_summary}
\end{figure}

We identify four major limitations in existing AD solutions in popular frameworks~\cite{adifor,jax,zygote,enzymead,pytorch}.
\textbf{First}, each tool supports a limited set of languages that do not cover the domain scientists' needs and often require the maintenance of multiple tools for different parts of the same project. For example, JAX~\cite{jax} and PyTorch~\cite{pytorch} work on Python only, while Zygote~\cite{zygote} works on Julia~\cite{julia} code. \textbf{Second}, these tools require domain scientists to modify their code to match the specific requirements of their AD frameworks, such as the immutability of arrays in JAX~\cite{jax}. \textbf{Third}, performance optimization is delegated to domain scientists. \textbf{Lastly}, a fundamental challenge in reverse-mode AD is the re-materialization problem~\cite{eval_derivatives}, which involves balancing memory usage and recomputation. By default, most frameworks store all intermediate values from the forward pass necessary for gradient computation and offload the recomputation decision to domain scientists~\cite{user_driven_check}, leading to substantial memory usage in large-scale applications~\cite{revolve799}. Due to these limitations, domain experts manually write the derivatives code for scientific applications. This increases code complexity, reduces performance, and poses substantial maintainability challenges, limiting AD's applicability in scientific domains~\cite{tapenade}.
%
To the best of our knowledge, no state-of-the-art AD tool solves all these problems efficiently (Table~\ref{tab:related_works}).

This work presents DaCe AD, a general, efficient automatic differentiation engine that requires zero code rewrites while offering 4.1x (geo-mean) speedup on average over state-of-the-art AD system (Figure~\ref{fig:dacead_summary}).
We use symbolic reverse-mode automatic differentiation and a novel ILP-based automatic checkpointing technique to compute gradients efficiently. Our solution is implemented on top of DaCe~\cite{dace}
an DaCeML (Section~\ref{sec:background}), which supports porting programs written in Python, PyTorch, ONNX~\cite{onnx}, and Fortran into the Stateful DataFlow multiGraph (SDFG) IR for AD. This data-centric intermediate representation is well-suited for AD, as it facilitates tracking data movement and dataflow analysis necessary for efficiently addressing traditional AD challenges, such as managing data overwrites, flowing gradients through loops, and storing/recomputing intermediaries.

\begin{table}
    \centering
    \includegraphics[width=1\linewidth]{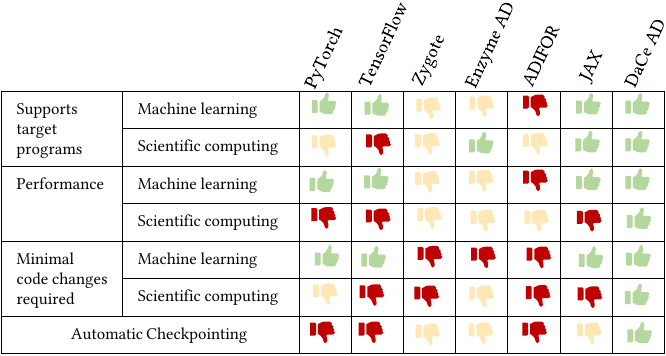}
    \caption{Overview of existing solution for automatic differentiation.}
    \label{tab:related_works}
\end{table}

We start by integrating the concept of a \emph{Critical Computation Subgraph (CCS)} into DaCe's SDFGs (Section~\ref{dacead_bwd}).
With the subgraphs, we can create the backward pass and apply
automatic differentiation to sequential and parallel loops (Section~\ref{ad_for_loops}).
Finally, we optimize the selection of an ideal number of stored arrays by formulating it as an ILP problem (Section~\ref{sec:tradeoffs}).

We evaluate our solution on NPBench~\cite{npbench}, a benchmark suite designed for numerical and scientific workloads in HPC. It contains a wide range of programs, from weather stencils to full deep learning models. Our results, presented in Section \ref{evaluation}, demonstrate an average speedup of over $92\times$ for gradient calculation versus JAX JIT on this benchmark suite. 
The main contributions of this work are as follows:

\begin{itemize}
  \item An automatic differentiation framework that supports both machine learning models and scientific computing codes within a unified environment, facilitating their integration. 
  \item A novel ILP-based solution to the re-materialization problem for reverse-mode AD~\cite{eval_derivatives}.
  \item An easy-to-use AD solution that outperforms JAX by over $92\times$ on average for gradient calculation on NPBench.
     
\end{itemize}

\label{sec:background}

\begin{figure}[t]
    \centering
    \includegraphics[width=1\linewidth]{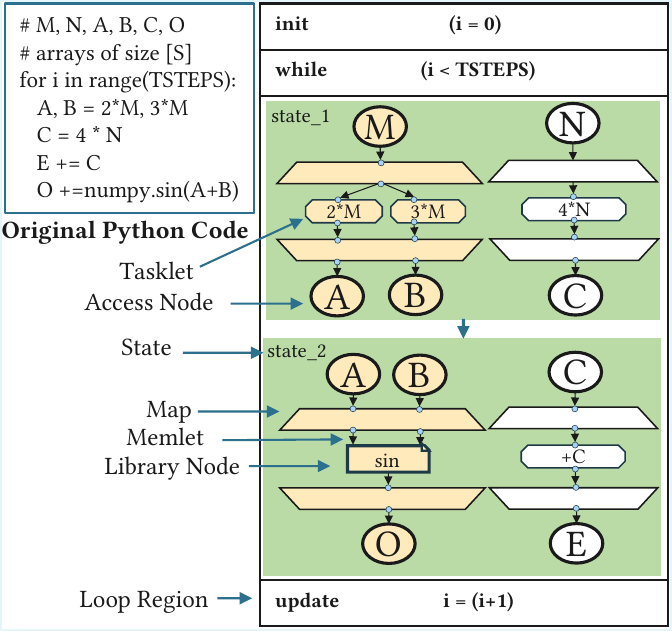}
    \caption{Example of an SDFG before optimization. Elements in yellow represent the elements of the program required in the backward pass to flow gradients through starting from the output \textit{O}. This represents the critical computation subgraph discussed in Section \ref{identifying_ccs}.}
\label{fig:background_dace}

\end{figure}

DaCe~\cite{dace} is a parallel programming framework that translates Python/NumPy~\cite{numpy} code and other languages, such as Fortran, into high-performance CPU, GPU, and FPGA implementations. DaCeML~\cite{daceml} is a DaCe extension that provides a PyTorch/ONNX frontend to generate SDFGs with basic AD capabilities. Internally, both frameworks use the Stateful DataFlow multiGraph (SDFG). Figure \ref{fig:background_dace} illustrates an SDFG, with its components defined below:
\begin{itemize}
    \item \textit{Access Node.} Oval-shaped nodes in the SDFG that provide access to data containers. Outgoing edges represent reads, while incoming edges indicate writes to the referenced data container.
    \item \textit{Memlet.}
Memlets represent data movement in the SDFG, specifying the data subset, amount of data transferred, etc.
    \item \textit{Tasklet.}
Tasklets represent fine-grained computations in the SDFG.
    \item \textit{Map.} A Map is represented by an entry (trapezoid) and exit (inverted trapezoid) nodes, defining a parallel region executed multiple times based on its iteration space. 
    \item \textit{Library Node.}
Library Nodes are specialized nodes that represent functions expanded during compilation into efficient library calls or native SDFG representations.
\item \textit{State.}
States contain a combination of previously defined nodes, and represent distinct stages of execution -- similar to steps in state machines or basic blocks in control flow graphs.
\item \textit{Loop Region.}
A construct in SDFGs representing sequential loops, where the loop's body executes a parametric number of times based on a conditional expression. The loop in Figure \ref{fig:background_dace}, for example, is a sequential timestep loop that iterates from \textit{0} to \textit{TSTEPS}.

\end{itemize}

 

DaCeML provides AD by implementing symbolic automatic differentiation~\cite{daceml}. This combines the precision of symbolic differentiation~\cite{art_ad} with the efficiency of automatic differentiation. It differentiates fine-grained computations and combines the output using the chain rule~\cite{calculus}. However, it only supports SDFGs with a single state and does not allow mutating arrays. It stores all values that need to be forwarded to the backward pass and does not provide recomputation options. 

These restrictions limit its applicability to scientific computing programs. Sequential loops and programs with control flow cannot be expressed in a single state. Our solution extends the symbolic automatic differentiation module to remove the above limitations. Furthermore, we add the possibility of recomputing arrays in the backward pass and an automatic solution for the re-materialization problem.

\section{The Critical Computation Subgraph -- The Key to the Backward Pass}
\label{dacead_bwd}
Creating the backward pass in DaCe AD translates to \emph{reversing} an SDFG and creating its corresponding \emph{backward} SDFG, which computes the gradients of an output (dependent variable) with respect to a set of input arrays (independent variables). This task is an excellent fit for DaCe's data-centric representation, which facilitates the most important challenge in AD: tracking dataflow. 

We identify three steps required to analyze the backward pass dataflow.
\begin{enumerate}
    \item Calculate the minimal subgraph (henceforth called critical computation subgraph, or CCS) that contains \textit{only} those computations that describe how the independent variables contribute to the dependent variables.
    \item For each element in the CCS, reverse the respective SDFG element in isolation.
    \item Connect the reversed elements using a straightforward, classical AD approach~\cite{eval_derivatives}.
\end{enumerate}

The engineering details referring to the reversal of most SDFG elements, as well as their connection, can be found in the documentation of our tool, which is publicly available \footnote{\url{https://github.com/spcl/dace/tree/dace_ad}}.
In the rest of this section, we focus on creating the critical computation subgraph (CCS) itself.


\paragraph*{The Critical Computation Subgraph}

\label{identifying_ccs}

\begin{figure}[t]
    \centering
    \begin{subfigure}[b]{0.48\textwidth}
        \centering
        \includegraphics[width=\textwidth]{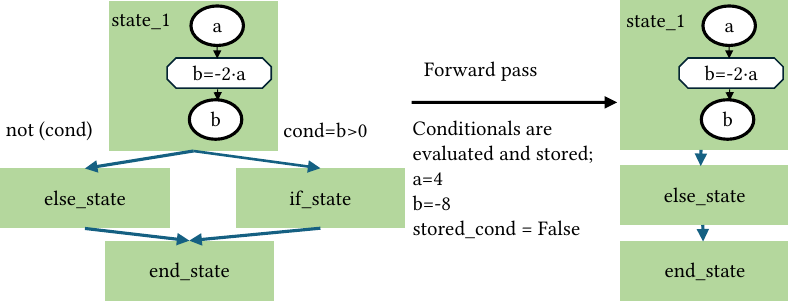}
        \caption{Forward pass evaluation stores results of conditional evaluations.}
        \label{fig:cf_fwd}
    \end{subfigure}
    \hfill
    \begin{subfigure}[b]{0.48\textwidth}
        \centering
        \includegraphics[width=\textwidth]{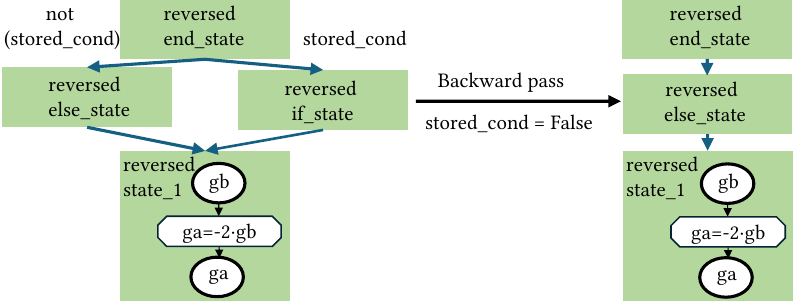}
        \caption{Stored conditionals values are used in the backward pass to remove unreached states.}
        \label{fig:cf_bwd}
    \end{subfigure}
    \caption{Backward pass gradient propagation through control flow.}
    \label{fig:store_maps}
\end{figure}

To identify the CCS, we perform a reverse breadth-first traversal~\cite{intro_algo}, starting from the dependent variable (output), and each explored node is added to the critical computation subgraph (CCS). The reverse breadth-first exploration represents all the elements that contribute to the dependent variable. 

We apply this algorithm to the two states in Figure \ref{fig:background_dace} independently from the loop. We are interested in retrieving the gradients of the output \textit{O} with respect to the input array \textit{M}. All the elements and operations contributing to the output \textit{O} are colored in yellow. To obtain this subgraph, we explore the graph in breadth-first order starting from the data node \textit{O}. Crucially, we must ensure the reverse breadth-first exploration propagates through states. This is necessary to track globally what data contributes to the output and how. 

In \textit{state\_1} of Figure \ref{fig:background_dace}, we start the reversed BFS exploration from all occurrences of \textit{A}, \textit{B}, and \textit{O}. This is because we know that \textit{A} and \textit{B} contribute to \textit{O}, so any operation that contributes to these two arrays will indirectly contribute to the dependent variable and thus should be tracked in the CCS. 

\paragraph{SDFGs with Control Flow}
In the case of control flow between multiple states, the BFS exploration will return elements that \textit{potentially} contribute to the output. This overestimation arises because the exact CCS depends on runtime-evaluated conditions, which are unavailable at compile time. Having an overestimation of this critical graph does not affect the correctness of the gradients since the additional states are within conditionals in the backward pass, and we prune this subgraph at runtime to match the CCS once the conditions of the forward pass are evaluated. An example of this pruning at runtime is shown in Figure \ref{fig:cf_fwd}. Similarly, the backward SDFG will be pruned at runtime to flow the gradients through the states executed (Figure~\ref{fig:cf_bwd}).

\begin{figure}[t]
    \centering
    \includegraphics[width=1\linewidth]{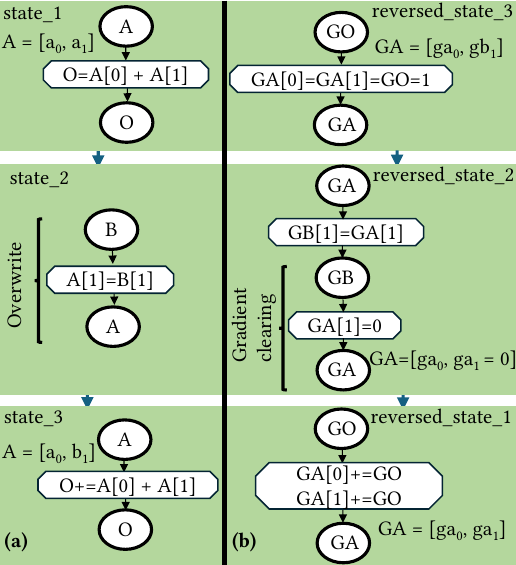}
    \caption{Example of gradient accumulation in the forward SDFG (a) and clearing in the backward SDFG (b).}
    \label{fig:grad_accu}
\end{figure}

\paragraph{Gradient Accumulation}
\label{gradient_accumelation}
Any array read in the forward graph will result in a write in the backward graph, because any read operation in the forward graph represents a contribution to the output. Since an array can contribute in many ways to the output, we will need to accumulate gradients for arrays that are read multiple times in the forward pass to account for the different contributions. We initialize all gradient arrays to zero at the time of their allocation and always accumulate gradients.

Because we allow overwriting, and since overwrites signify that we are now tracking a new set of values that so far have not contributed yet to the output, we need to reinitialize all gradient indices for the overwritten values back to zero. Figure~\ref{fig:grad_accu} shows an example where this gradient clearing and accumulation is necessary. When we reverse \textit{state\_2} and detect the overwrite, we add the gradient clearing Tasklet in \textit{state\_2\_reversed}. This guarantees that the contribution of $b_1$ is not erroneously counted toward \textit{A}.

\section{Compiling Efficient AD Loops Quickly}
\label{ad_for_loops}
Handling loops efficiently has been a persistent challenge for AD tools~\cite{source_loop,ad_parallel_loops}.
In this Section, we show how we compactly flow gradients through loops in the forward pass without unrolling them completely. This is important both for achieving good performance and for reducing the compilation time when applying AD to large loops.
First, we will present a classification of loops that benefit the most from our technique; then, we will show how we reverse both sequential and parallel loops at the SDFG level.

\subsection{AD Loop Taxonomy}
\label{taxonomy}
 We support differentiating programs that combine assignments, branching, parallel loops (SDFG Maps), and arbitrary nests of sequential for-loops that iterate over a structured index set without \texttt{break} or \texttt{continue} statements. This ensures a static iteration space and predictable control flow for gradient computation.
 The general form of this class of loops is:

 \texttt{for (int i = L; i < U; i = i + s) 
 \{
 ...
 \}
 } 
 
\begin{figure}
    \centering
    \includegraphics[width=1\linewidth]{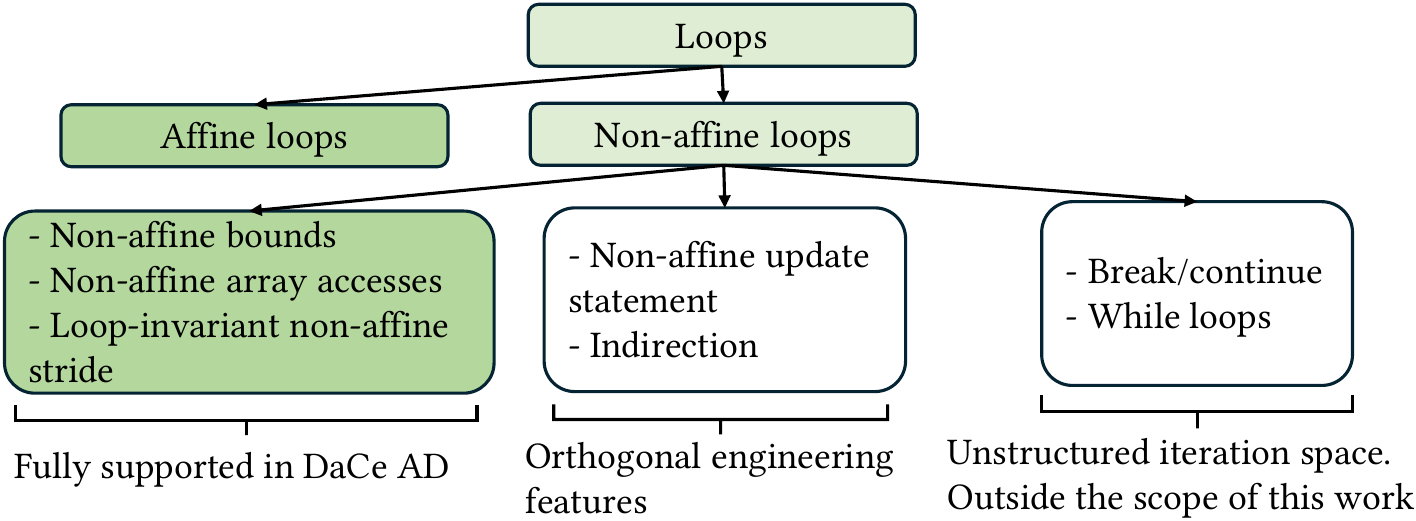}
    \caption{Taxonomy of loops for automatic differentiation.}
    \vspace{-0.6cm}\label{fig:tax}
\end{figure}

 In the loop header above, \textit{L}, \textit{U}, and \textit{s} are positive or negative integers that can be the result of affine and non-affine functions of loop-invariant parameters or outer loop iterators. The loop body may not contain instructions affecting the loop header. Resulting from these conditions, we offer a classification of loops we support in DaCe AD (colored in green) in Figure \ref{fig:tax}. 
 
 The methodology presented in Section~\ref{sequential_loop_rev} can be extended to support while-loops and break/continue in for loops by tracking exactly which iterations were executed in the forward pass and applying the reversal procedure
 on this set of iterations. However, the strength of our approach is the ability to generate compact loops in the backward pass automatically. Because of their unstructured iteration space, while-loops and for-loops with breaks cannot result in compact loops in the backward pass~\cite{art_ad} and, thus, are outside the scope of our work. 
 
 We can extend non-affine loop support to any non-affine update statement function $f$ that has a defined inverse function $f^{-1}$ by symbolically extracting $f^{-1}$ for the backward loop updates. We fully support non-affine stride values since these are stored as \textit{values} in memory and reused in the backward pass regardless of the invertibility of the function that generates the non-affine strides.  

\paragraph{Limitations.}
 Since we focus on reverse-mode AD, our approach computes the gradient with respect to a single program output. This approach is generalizable, as the process can be trivially repeated for each output of interest. We implement our solution in DaCe and therefore operate on programs currently representable by the framework\footnote{\href{https://spcldace.readthedocs.io/en/latest/frontend/pysupport.html}{spcldace.readthedocs.io/en/latest/frontend/pysupport.html}} - this means not supporting recursion and arbitrary Pythonic lists that do not iterate over index sets. Our current implementation focuses on analyzing and supporting different loop types. Two features beyond the scope of this work are indirections and complex number operations~\cite{complex_ad}, but these are engineering efforts rather than fundamental limitations of the approach.

 On NPBench, we support gradient calculation for $82\%$ ($38/46$ AD-compatible programs) of the benchmarks without code changes, thus showing that the restrictions described in this paragraph do not significantly limit the applicability of our approach. 

\subsection{AD of Sequential Loops in DaCe}
\label{sequential_loop_rev}

\begin{figure}[t]
    \centering
    \includegraphics[width=1\linewidth]{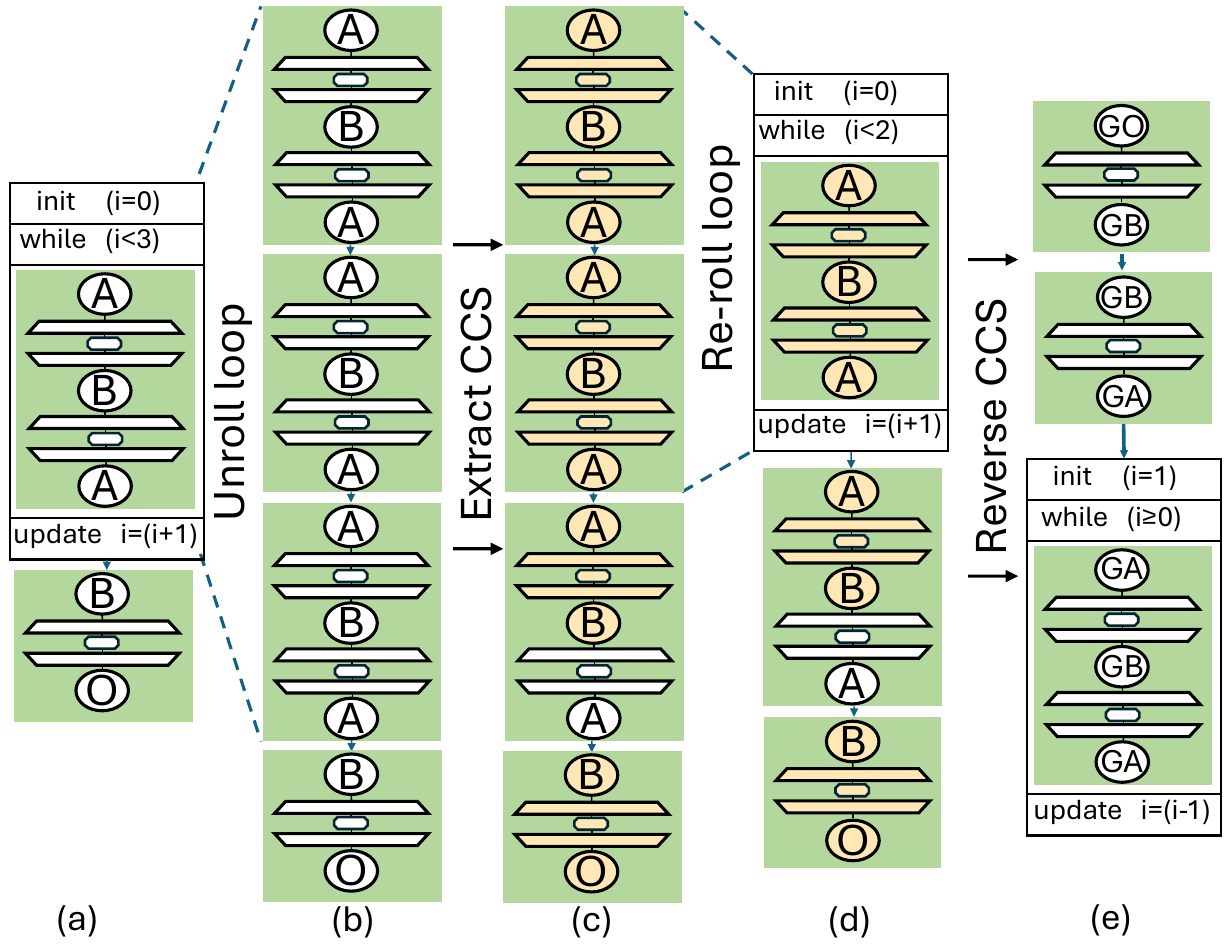}
    \caption{CCS extraction and reversal for a loop. (a) The initial program SDFG. (b) The SDFG after unrolling the loop. (c) The CCS, colored in yellow. (d) The SDFG after rerolling the matching CCS iterations. (e) The backward SDFG.}
    \vspace{-0.6cm}\label{fig:loop_reversal_example}
\end{figure}

To differentiate sequential loops, we start by first identifying the CCS. Sequential loops complicate the identification of the critical computation subgraph for AD, as the contributions of each iteration to the output may vary. A forward loop body does not always result in a \textit{single} critical computation graph that can be reversed and used as a loop body in the backward pass. Unrolling the for loop and applying the BFS intersection algorithm described in Section \ref{identifying_ccs} may yield non-identical critical subgraph views for each iteration's body. 

Let \textit{L} be a loop of \textit{$N > 0$} iterations (from 0 to \textit{$N-1$}) through which we are interested in flowing gradients. If we unroll \textit{L}, we will create \textit{$N$} sequential replications of the loop body, each representing a single iteration of \textit{L}. Figure~\ref{fig:loop_reversal_example}\hyperref[fig:loop_reversal_example]{a} presents an example of a program with a sequential for loop. We are interested in getting the gradients of \textit{O} with respect to the input \textit{A}. Figure \ref{fig:loop_reversal_example}\hyperref[fig:loop_reversal_example]{b} shows the SDFG after completely unrolling the loop. Applying the procedure presented at the start of Section \ref{identifying_ccs}, we iterate backward in the SDFG to track what elements contribute to \textit{O}. The CCS resulting from this step is visualized as yellow elements in \ref{fig:loop_reversal_example}\hyperref[fig:loop_reversal_example]{c}. Because we need to track what directly contributes to the output \textbf{and} what contributes to those contributions, we will potentially start the reverse BFS search with a different set of starting point nodes for each iteration's body. 

It is possible that after exploring a sufficient number of loop iteration bodies, the set of starting point nodes contains all the required nodes and will not change for future iterations after this point. Since the loop body is identical across iterations, the CCS remains consistent for all subsequent iterations. In this case, we can reroll the loop and then reverse its CCS in the backward pass. In Figure~\ref{fig:loop_reversal_example}\hyperref[fig:loop_reversal_example]{c}, we notice that after exploring the body of the last iteration of the loop, we add both arrays \textit{A} and \textit{B} to the set of elements that contribute to \textit{O} and should be tracked, which also serves as the starting point of the reverse BFS for the next states. After this state, and because our starting point for the reverse BFS contains all the SDFG arrays, all of the next CCS states match, and we can first reroll the loop (Figure \ref{fig:loop_reversal_example}\hyperref[fig:loop_reversal_example]{d}) and then reverse it in the backward pass (Figure \ref{fig:loop_reversal_example}\hyperref[fig:loop_reversal_example]{e}). 

\begin{figure}
     \centering
     \includegraphics[width=1\linewidth]{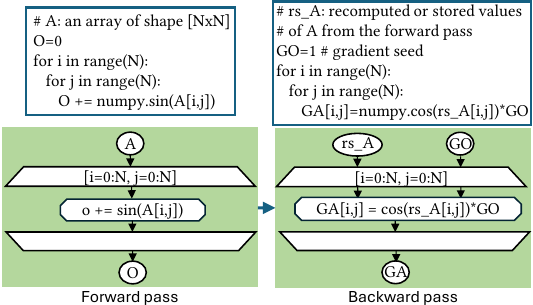}
     \caption{Example of automatic differentiation through parallel loops (SDFG Maps).}
     \vspace{-0.6cm}\label{fig:mapad}
 \end{figure}
 
While the example above showcases the concept, we do not unroll the loop and reroll it in practice. Instead, we apply a dataflow analysis pass to discover the loop CCS. This means we do not require the loop to be unrollable to reverse it compactly. 


\subsection{AD of Parallel Loops (Maps) in DaCe}
\label{parallel_loop_rev}
 A Map represents a parallel index loop over an $N$-dimensional index set. For example, the map in Figure~\ref{fig:mapad} is a 2-dimensional map that iterates over all the elements of an \textit{[NxN]} array and applies \texttt{np.sin()} to each element, then adds it to a scalar output \textit{O}. Because of their structured iteration space, maps are easier to reverse. 
 
 If a map is within the CCS of the forward SDFG, a new Map is created in the backward pass with the same range, as depicted in Figure~\ref{fig:mapad}. We additionally reverse the Tasklets and any other SDFG elements (Memlets, Maps, etc.) within the map to ensure that gradients flow through the parallel region.


\section{Store-recompute trade-off}
\label{sec:tradeoffs}

\begin{figure}[t]
    \centering
    \begin{subfigure}[b]{0.48\textwidth}
        \centering
        \includegraphics[width=\textwidth]{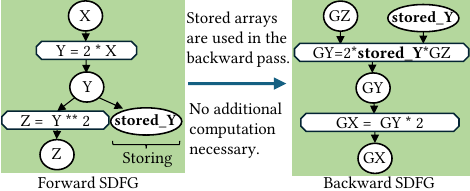}
        \caption{Storing arrays for the backward pass.}
        \label{fig:store}
    \end{subfigure}
    \hfill
    \begin{subfigure}[b]{0.48\textwidth}
        \centering
        \includegraphics[width=\textwidth]{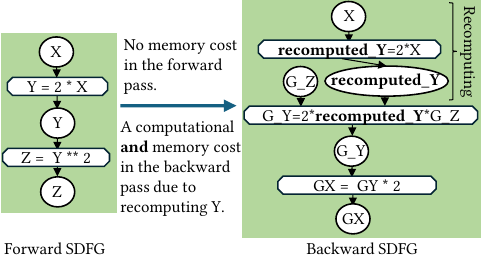}
        \caption{Recomputing arrays in the backward pass.}
        \label{fig:recomp}
    \end{subfigure}
    \caption{Example of the trade-off between store (a) and recomputation (b) in non-linear operations.}
    \vspace{-0.6cm}\label{fig:store-recompute-example}
\end{figure}

Non-linear operations in the program will require passing/forwarding data constructed in the forward pass to the backward pass~\cite{eval_derivatives}. Figure~\ref{fig:store-recompute-example} shows an example of the trade-off between (~\ref{fig:store}) storing the value of $Y$  or (~\ref{fig:recomp}) recomputing/re-materializing it in the backward pass.
While we can save computations by storing values throughout the forward pass to be later used in the backward pass, it can incur significant memory costs. 

Recomputation trades off compute time to avoid storing values in memory. The problem of identifying what to store and what to recompute is known as the re-materialization problem~\cite{art_ad}. By default, DaCe AD stores intermediate values. While we offer the possibility for a user to manually decide to recompute specific arrays, our approach goes further by also introducing an automatic solution for this problem. 

A core part of our contribution is a novel algorithm that models this question as an Integer Linear Programming (ILP)~\cite{ilp} problem, where a binary decision variable in the system represents whether to store or recompute each target array. 

Given a user-defined memory constraint in MiB, we model the memory usage of the program as a sequence of memory allocation and deallocations. We identify where the allocations and deallocations of each option (storing or recomputing) would be inserted within the memory sequence, and we introduce parametric elements that are a function of the size of the arrays that need to be forwarded. We want each sequence element (representing a pre-calculated memory measurement) to be less than the user-defined memory limit, which will serve as the set of constraints for the ILP. We model the computational cost of recomputing each array and minimize the cost under these constraints. This translates to computing the gradients as fast as possible while respecting the memory constraint.

\subsection{Motivating example}
\label{ilp_motivating_example}
Listing~\ref{lst:ilp_example} introduces an example that will demonstrate the memory measurement terms and the structure of the ILP system.

\begin{lstlisting}[
  belowskip=-1.5\baselineskip,
  language=Python, 
  caption={Re-materialization ILP example.}, 
  label={lst:ilp_example}, 
  float,
  frame=none,
  commentstyle=\color{green!50!black},
  keywordstyle=\color{blue},
  stringstyle=\color{red},
  basicstyle=\ttfamily\footnotesize,
  breaklines=true,
  breakatwhitespace=true,
  showstringspaces=false,
  alsoletter={.},
  numbers=left,
  numberstyle=\tiny\color{gray},
  numbersep=5pt,
  escapeinside={(*@}{@*)},
  alsoletter={.},
  morekeywords={np, dace, float32},
  emphstyle={\color{purple}},
  emph={[2]np.multiply, np.sin, np.sum},
  emphstyle={[2]\color{cyan!60!black}},
  emph={[3]dace},
  emphstyle={[3]\color{magenta!80!black}},
  emph={[4]dace.float32, @dace.program},
  emphstyle={[4]{\color{magenta!80!black}\color{purple}}},
]
@dace.program
def foo(C: dace.float32[N, N], D: dace.float32[N, N]):
    A0 = np.multiply(C, D)  
    sin0 = np.sin(A0) # Non-linear operation. 
    D = D * 6
    A1 = np.multiply(C, D)
    sin1 = np.sin(A1) 
    D = D * 3
    A2 = np.multiply(C, D)
    sin2 = np.sin(A2)
    return np.sum(sin0 + sin1 + sin2)
\end{lstlisting}

\paragraph{Storing and Recomputation Costs}
Only arrays $\{A_0, A_1, A_2\}$ are used in the non-linear operation \texttt{np.sin()} and thus need to be forwarded to the backward pass. For $N=3620$, the size $S_i$ of each array $A_i$ in the program is $50$ MiB, including the arrays to be forwarded ($i \in \{1,2,3\}$):

$S_0 = S_1 = S_2 = 50$ MiB.

The further the computation is down the forward pass dependency graph, the more costly it will be (both in terms of memory and computation) to recompute it in the backward pass.

We use the number of floating point operations to estimate the recomputation cost of each array. Recomputing array $A_0$ will require a single element-wise multiplication of two $[NxN]$ arrays. This translates to a recomputation cost $c_0\approx13$ MFLOPS for $N=3620$. Recomputing array $A_0$ will not require additional memory given the global inputs. In contrast, recomputing $A_1$ will require the multiplication of the input $D$ by $6$ performed in Listing~\ref{lst:ilp_example}, line 5, which results in an intermediate array of size $50$ MiB, then generating $A_1$. Recomputing $A_2$ will require two element-wise multiplications, and thus $c_1=2*c_0\approx26$ MFLOPS.

Applying these rules, we can determine the peak memory overhead $R_i$ for recomputing each array $A_i$, $i \in \{1,2,3\}$:

$R_0 = 0, R_1 = 50, R_2 = 100$ MiB.

Similarly, we estimate the computational cost $c_i$ (in MFLOPS) of recomputing each of the arrays $A_i$, $i \in \{1,2,3\}$:

$c_0 \approx 13, c_1 \approx 26, c_2 \approx 39$ MFLOPS.

It is apparent that recomputing $A_2$ will incur a higher computational cost ($c_2$) and memory overhead ($R_2$) since it would require running the entire forward pass to generate the needed dependencies (array $D$). 

\paragraph{Decision Variables and Objective Function}
We define three binary decision variables $v_0, v_1, v_2$, one for each array to be forwarded. $v_0$ for example, decides whether array $A_0$ should be stored ($v_0 = 1$) or recomputed ($v_0 = 0$). 

The objective function to minimize will be the sum of the computational costs of recomputed arrays:
\begin{align*}
    f(v_0, v_1, v_2)&= c_0\cdot(1-v_0) + c_1\cdot(1-v_1) + c_2\cdot(1-v_2)\\
    f(v_0, v_1, v_2)&= 13\cdot(1-v_0) + 26\cdot(1-v_1) + 39\cdot(1-v_2)
\end{align*}
If $A_0$ is stored, $v_0 = 1$, and the term $c_0$ does not appear in the objective function. Minimizing this cost function translates to storing as much as possible ($v_i=1$) while respecting the memory constraint.

\paragraph{Memory Measurement Sequence}
Let $m_i$ be a sequence of total memory usage values for the program, each $m_i$ corresponding to either an array being allocated or freed. In essence, this sequence can be viewed as a timeline of how the peak memory usage evolves during program execution; some of these measurements will be a function of the decision variables $v_i$. Thus, we can model the peak memory measurement of the program as a function of what we decide to store and recompute. We identify points in the program where the storing and recomputation of an array $A_i$ could be introduced, and we add the memory cost of each strategy depending on the decision variable $v_i$. 

 An excerpt of an example of the memory measurement sequence is presented below.
\begin{align*}
    m_1 &= 50, \quad \text{(Allocation of program array)} \\
    m_2 &= 50 + v_0 \cdot 50, \quad \text{(variable for storing $A_0$)}\\
\end{align*}
Some of the measurements contain the decision variables -- this represents us having identified that if we were to store $A_0$ ($v_0 == 1$), then the allocation for the new array ($50$ MiBs) that contains these values will happen at time step $2$.

Because recomputation itself has a memory consumption overhead, we also need to identify the recomputation time stamps in the backward pass and model them in our constraints:
\begin{align*}
    m_{21} &= m_{20} + (1-v_1) \cdot R_1 + S_1 \cdot (1-v_1), \\
    m_{22} &= m_{21} - (1-v_1) \cdot R_1, \\
\end{align*}
The term $m_{21}$ means that if we decide to recompute the array $A_1$ ($v_1 = 0$), then we will have a memory overhead of $R_1$ from the recomputation and an allocation of $S_1 = 50$ (MiB) will be required to host the new values. The term $m_{22}$ shows the immediate removal of the recomputation memory overhead. 

\paragraph{ILP Problem}
We introduced an auxiliary variable, $t$, to represent the peak memory usage of the program for a given store-recompute configuration. We want to minimize the function $f$ subject to the constraints:
\begin{align*}
t &\geq m_i \quad \text{for all } i = 1, 2, \ldots, k \\
t &\leq 500  \quad \text{(Our memory constraint in this case is 500  MiB)}\\
v_i &\in \{0, 1\} \quad \text{for all } i = 1, 2, 3
\end{align*}

\paragraph{ILP Solution}
In this example, the ILP solver must recompute only one array to meet the constraints. Because the computational cost of re-materializing $A_0$ is the least expensive and its recomputation block also consumes less memory, the solver chooses to store $A_1$ and $A_2$ and recompute $A_0$ since it has the least computational cost. Our tool implements a generalized ILP formulation similar to the approach summarized above, and we evaluate this program with DaCe AD in Section \ref{eval_ilp}.



\subsection{ILP-Checkpointing for SDFGs with Control Flow}

\begin{figure}[t]
    \centering
    \begin{subfigure}[b]{0.49\linewidth}
        \centering
        \begin{lstlisting}[
            language=Python, 
            label={lst:mem_example}, 
            frame=none,
            commentstyle=\color{green!50!black},
            keywordstyle=\color{blue},
            stringstyle=\color{red},
            basicstyle=\ttfamily\footnotesize,
            breaklines=true,
            breakatwhitespace=true,
            showstringspaces=false,
            alsoletter={.},
            % numbers=left,
            numberstyle=\tiny\color{gray},
            numbersep=5pt,
            escapeinside={(*@}{@*)},
            alsoletter={.},
            morekeywords={np, dace, float32},
            emphstyle={\color{purple}},
            emph={[2]np.multiply, np.sin, np.sum},
            emphstyle={[2]\color{cyan!60!black}},
            emph={[3]dace},
            emphstyle={[3]\color{magenta!80!black}},
            emph={[4]dace.float32, @dace.program},
            emphstyle={[4]{\color{magenta!80!black}\color{purple}}},
        ]
import numpy as np
rand = np.random.random 
N=3620
A = rand([N,N])    # state_1
B = np.ones([N,N]) # state_2
if A[0,0]>0: # go to state_3
    C = A * 2
    D = B * 4
else:        # go to state_4
    C = (A + B) * 2 
    D = C * 3
        \end{lstlisting}
        \caption{Python code.}
    \end{subfigure}
    \hfill
    \begin{subfigure}[b]{0.49\linewidth}
        \centering
        \includegraphics[width=\linewidth]{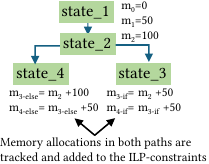}
        \caption{Forward SDFG.}
        \label{fig:ilp_control}
    \end{subfigure}
    \caption{Memory measurement sequence with control flow.}
    \vspace{-0.7cm}\label{fig:ilp_combined}
\end{figure}

The memory constraint must be respected regardless of which execution path is taken in a program with control flow. Thus, we model the memory measurement sequence for each path and add all of these as constraints to the final ILP. An example of the memory measurement sequence of a program with control flow can be visualized in Figure \ref{fig:ilp_control}. 

In this example, we add the recomputation costs of all the arrays (even those within conditionals) to the objective function to minimize.
Once we encounter the branching in \textit{state 2}, we save the last measurement from this state ($m_2$), and for each branch, we start a new measurement sequence that simulates the memory usage of running this branch. We additionally use preprocessing transformations in DaCe before applying AD to reduce the unnecessary control flow in programs. This is particularly useful in large weather simulations, such as ICON~\cite{icon}, where much of the control flow is used to choose which model configuration is used and can be removed when executing a specific configuration.   

\section{Evaluation}
\label{evaluation}

In this Section, we compare the performance and applicability of our solution with JAX~\cite{jax}, a state-of-the-art Python framework developed for both high-performance numerical computing and large-scale machine learning. One of the major features of JAX is that it offers gradient computation natively. 
We focus our comparison on NPBench~\cite{npbench}, a benchmark suite designed to cover various scientific domains such as machine learning, weather modeling, computational fluid dynamics, and quantum transport simulation.

We conduct experiments on a system with two Intel® Xeon® Gold 6154 CPUs (18 cores per CPU, 72 logical processors) running at 3.00 GHz (max 3.7 GHz). This x86\_64 architecture supports AVX-512, FMA, and SSE4.2 instructions. The system has two NUMA nodes, with a cache hierarchy of 32 KB L1, 1 MB L2 per core, and 25 MB L3 per CPU.

\subsection{Performance on NPBench}
\label{npbench_performance}
From NPBench, we exclude 14 benchmarks from the original 52 kernels for their incompatibility with AD. 
We remove as follows:
\begin{itemize}
    \item \textit{stockham fft}, \textit{scattering self energies}, \textit{contour integral}, \textit{mandelbrot one and two} -- they contain operations on complex numbers, which are outside the scope of this work.
    \item \textit{aziment naive}, \textit{nbody} \textit{aziment hist}, \textit{crc16}, \textit{floyd warshall}, and \textit{nussinov} -- they contain many points of discontinuity that cancel or make the gradients undefined
    \item \textit{spmv} -- it contains indirection
    \item \textit{channel flow} -- it contains a while loop
    \item \textit{cholesky2} -- it contains an external library call
\end{itemize} 

To be able to run reverse-mode automatic differentiation, we add a sum reduction on one of the output arrays chosen randomly for each program. As described in NPBench, all benchmarks work on \textit{np.float64} arrays except for the five deep learning benchmarks: \textit{mlp, lenet, resnet, softmax and conv2d} which use \textit{np.float32}. We use the same sizes for the inputs as the original NPBench paper, which are named "paper" size in the NPBench repository.
\begin{figure}
    \centering
    \includegraphics[width=1\linewidth]{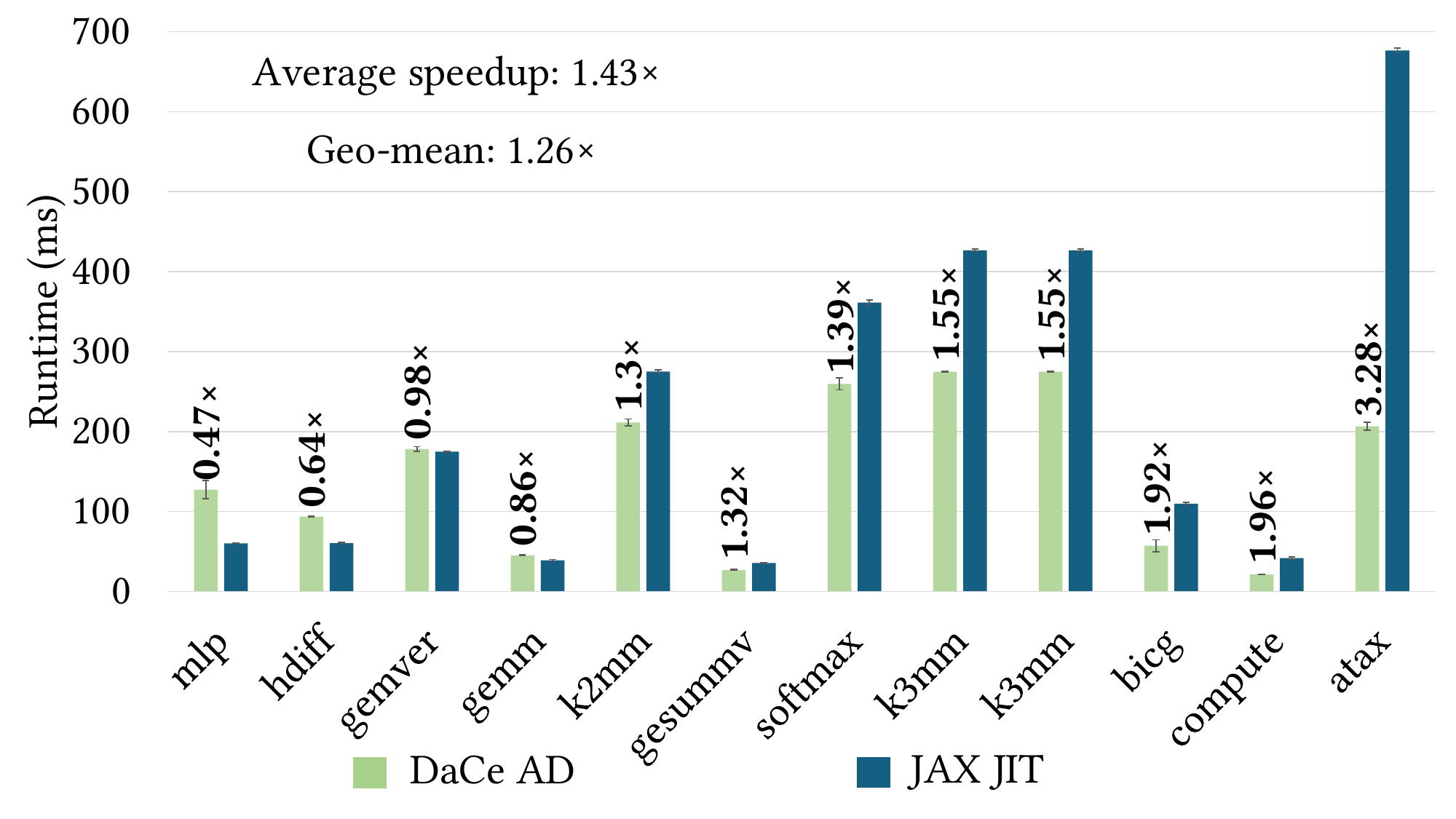}
    \caption{DaCe AD vs. JAX JIT: Performance on vectorized benchmarks. The data labels represent the DaCe AD speedup over JAX JIT.}
    \vspace{-0.7cm}\label{fig:eval_vectorized}
\end{figure}
\begin{figure*}
        \centering
        \includegraphics[width=1\linewidth]{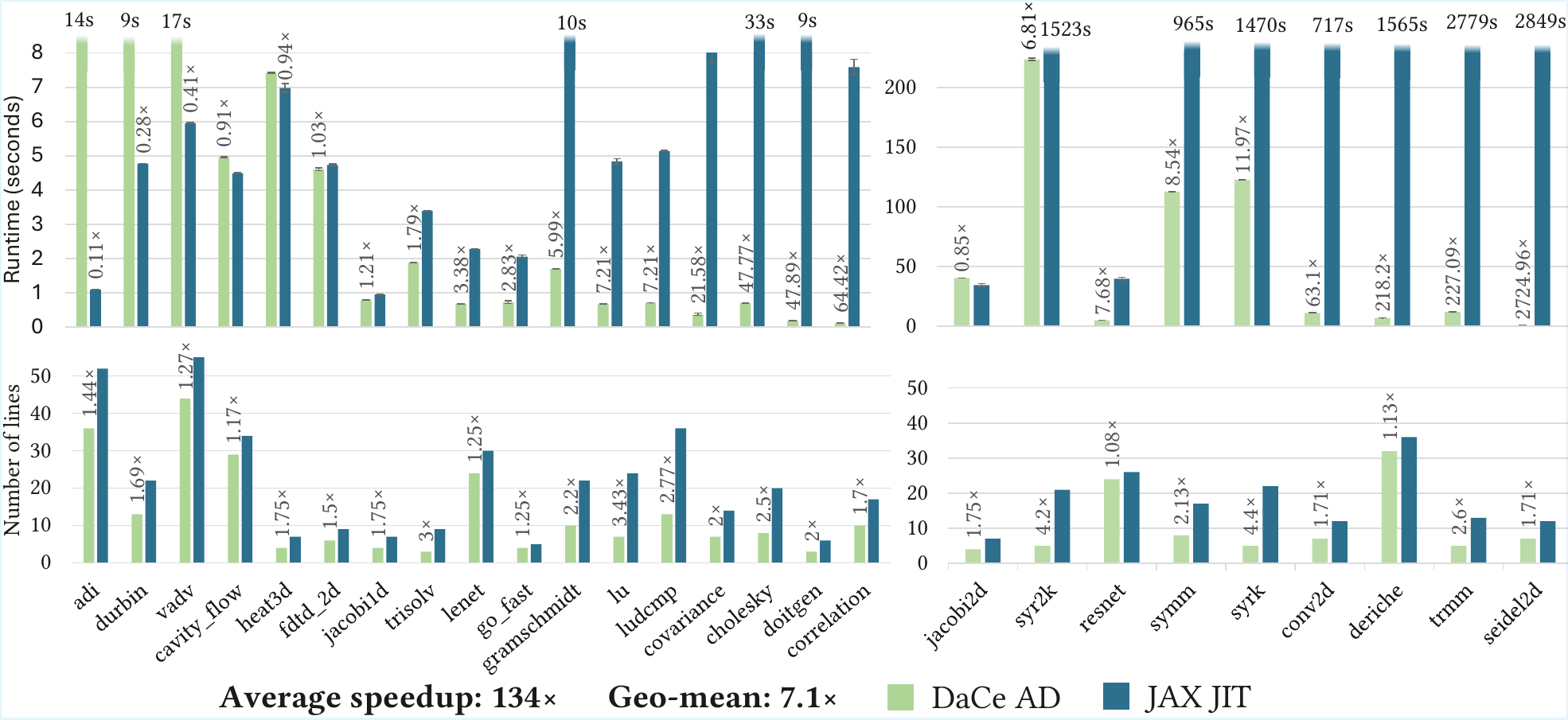}
        \caption{DaCe AD vs. JAX JIT - Non-vectorized benchmarks - Comparison of performance and forward-pass program size.}
        \label{fig:eval_non_vect}
\end{figure*}

We numerically validate all gradient outputs using the \textit{np.allclose} function. All runs for both JAX and DaCe AD use the store-all strategy. We apply post-AD optimizations through the DaCe automatic optimization pass. For both frameworks, a warmup run is added before measuring the execution time to avoid compilation overhead. Following performance evaluation guidelines~\cite{benchmarking}, each measurement is repeated 10 times to avoid noise, and CI bounds are reported. 
Overall, we outperform JAX on the remaining 38/52 benchmarks with an average speedup of $92\times$ and a geometric mean of $4.1\times$, which is calculated as the $n^{\text{th}}$ root of the product of the speedup ratios where $n=38$. We split these programs into two categories:

\subsubsection{Vectorized Programs} These programs do not contain any loops and apply operations on whole arrays or vectors. NPBench has twelve programs with either matrix-matrix or matrix-vector multiplication as the core operation. This is the target program for the JAX frameworks and the XLA compiler~\cite{jax} used in JAX JIT. Figure~\ref{fig:eval_vectorized} shows the performance of DaCe AD versus JAX JIT for vectorized programs. We outperform JAX with and without JIT on 8/12 programs with an average speedup of $1.43\times$ (geometric mean: $1.27\times$) when compared to JAX JIT.

The performance of programs in this category largely depends on having an optimized library call for matrix-matrix and matrix-vector multiplications and the input matrix sizes. In DaCe, we pattern-match through the backward SDFG to look for any \textit{matmul} operations that can be transformed into an optimized library call in the generated code. Furthermore, we can choose from multiple optimized libraries (intel-MKL~\cite{mkl}, CBLAS~\cite{cblas}, and cuBLAS~\cite{cublas}) depending on the target hardware and environment. Because we use JAX library calls through \texttt{jax.numpy}, JAX is also able to lower these calls to optimized implementations through XLA. 

\subsubsection{Non-vectorized programs}
Programs with loops, control flow, and individual array accesses are very common and important in scientific code, as demonstrated by their representation in NPBench (26/38 in our AD experiments). 
\paragraph{Necessary JAX Code changes}

To make a NumPy-style program compatible with JAX~JIT, every NumPy call needs to be replaced by its \texttt{jax.numpy} counterpart, and in-place updates such as \texttt{A[i] = 0} need to be rewritten as \texttt{A = A.at(i).set(0)} to satisfy JAX’s immutability rules.  
Loops must have static bounds (or be re-compiled for each bound) and are generally rewritten with \texttt{jax.lax.scan}, which supports automatic differentiation; inside the scan, dynamic indexing is handled through 
\texttt{lax.}\texttt{dynamic\_update\_slice}, and variable-length slices are accommodated by padding to a fixed shape and masking with \texttt{jnp.where}.
   
Figure~\ref{fig:eval_non_vect} compares the code size of the forward pass between JAX and DaCe AD. In all non-vectorized programs, the DaCe code is shorter than the JAX code. This is mainly because of the native support for the \texttt{range} loop in DaCe AD that requires no changes from the original NumPy code in contrast to the \texttt{jax.lax.scan}, which requires isolating the loop body as a pure function. Additionally, in cases like \textit{syr2k}, preparing masks for dynamic slicing adds additional lines on top of the loop redefinitions. 


\paragraph{Performance comparison on non-vectorized programs}

Even after rewriting to allow JAX JIT to optimize the gradient computation, we significantly outperform JAX on most benchmarks in this category (20/26) with an average speedup of $134\times$ (geometric mean $7.12\times$). There are three main reasons for this performance difference: First, JAX array slices translate to a \textit{lax.dynamic\_slice} call when the slice indices are not static, in contrast to DaCe slices, which are translated into direct memory accesses. The second performance hurdle for JAX is the immutability of arrays. A simple assignment to an index in the forward pass is translated into multiple lower-level operations in the backward pass that generate intermediate arrays (often proportional to the size of the input arrays). Finally, JAX adds bound check operations within the loop to propagate the slice gradients correctly and avoid illegal memory accesses. In contrast, DaCe checks the validity of the assignment at the SDFG Memlet level, which showcases the benefit of the data flow-centric representation. 

Each one of the three problems discussed above adds overhead at the iteration level, significantly slowing down large loops. The difference in speedup from one benchmark to another seen in Figure~\ref{fig:eval_non_vect} is due to the size of the iteration space of the loops and the operation within the loop. The larger the iteration space, the more JAX suffers from the overhead of these three constraints. Additionally, the difference between the speed up in \textit{syrk} ($\approx12\times$) and \textit{trmm} ($227\times$), which have a similar iteration space (the innermost loop is repeated $1,200,000$ times in both cases) is due to the dot operation within the \textit{trmm} loop which causes additional overhead (dynamic slicing of the inputs to the dot product). \textit{Gramschmidt}, for example, has a smaller iteration space first because the array sizes are smaller ([240, 200] in comparison to [400,400] in \textit{Seidel2d}) and because the iteration space is triangular, so we only iterate through a triangle in the matrix and not through all the values. There, JAX's performance is not that strongly affected by overheads of each iteration and we only outperform it by a factor of $\approx6\times$.

\begin{figure}
    \centering
    \includegraphics[width=1\linewidth]{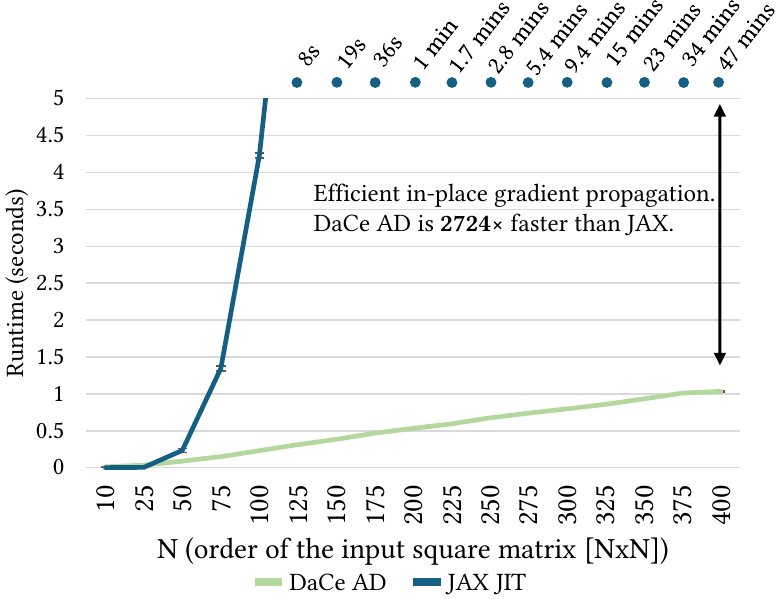}
    \caption{Variation of the size of the input 2D array for the \textit{Seidel2d} benchmark.}
    \vspace{-0.4cm}\label{fig:variations}
\end{figure}

\subsection{Case Study: Seidel2d}
\label{seidel2d_case_study}
In the following section, we conduct a deep analysis of the Seidel2D kernel\footnote{\href{https://github.com/spcl/npbench/blob/main/npbench/benchmarks/polybench/seidel_2d/seidel_2d_jax.py}{Siedel2d JAX JIT code}}, which is a stencil computation on a two-dimensional array within a time-step loop.
We will use it to explain how DaCe AD's performance can be over 2,700 times better than JAX's performance.

We focus on the innermost loop (\textit{loop3} in the linked code) only since that is the most computationally expensive part of the code. This particular loop contains a single assignment. 
First, two dynamic slices are added to obtain \texttt{A[i, j]} and \texttt{A[i,j-1]}. Each of these calls is preceded by index preparation, additional bound checks, and adjustments for negative indices. These bound checks are only present in the backward pass, making it significantly slower than the forward pass, which directly scatters the slice using the forward indices. In DaCe AD, we do the bound checking through symbolic analysis at the SDFG level, ensuring that assignments for each loop iteration are within bounds. 

This means that the generated DaCe code is free of any unnecessary bound checks, especially within the loops. We additionally remove the overhead of dynamic slicing in the backward pass because we do not require static shapes for the slices, which translates to cheap pointer movements in the DaCe-generated code. Towards the end of the JAX loop iteration, a new array is created carrying the gradients from this iteration.

Note that even though only a single array value is updated, a \textit{[N, N]} array is created at each iteration. In contrast, DaCe AD updates this through a single assignment to the gradient array, which translates to a single write in memory. For the size with which we run our experiment, this translates into creating a \textit{ [400,400]} array for each iteration. Since the loop is of depth 3, with time-step loop followed by the iteration over all the values of a 2D matrix, and with the number of timesteps TSTEPS of 100, this extra overhead is repeated: $nb\_iterations = TSTEPS*N^2=16,000,000$ times

\begin{figure}
        \centering
        \includegraphics[width=1\linewidth]{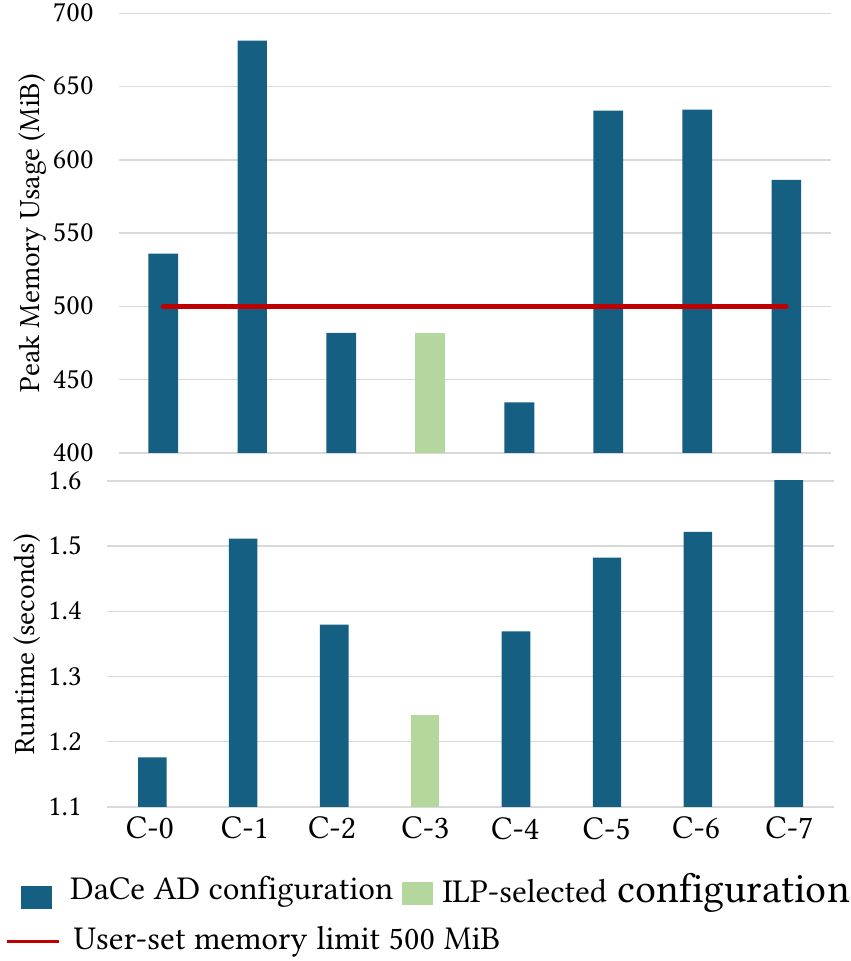}
        \caption{Performance and memory usage comparison of different store-recompute configurations. C-0, for example, represents storing all three arrays from the example presented in Section \ref{eval_ilp}.}
        \vspace{-0.5cm}
        \label{fig:eval_ilp_example}
\end{figure}

The combination of bound checking at the loop level, dynamic slicing, and creating large unnecessary arrays to respect immutability results in poor performance, allowing DaCe to be over $2,700\times$ faster than JAX. While JAX JIT obtains the gradients in over $47$ minutes, DaCe AD can compute the gradients in one second only. Figure~\ref{fig:variations} shows how the gradient calculation time evolves for different input sizes for both JAX JIT and DaCe AD on this benchmark. Because the dynamic slicing performance on larger arrays is slow and the complexity of the innermost loop is $O(TSTEPS\cdot N^2)$, increasing the size of the input array has a much bigger effect on the JAX JIT runtime. For smaller arrays ($N<50$), the overhead of dynamic slicing is negligible, and JAX JIT performs better than DaCe AD. If $N=10, TSTEPS=100$, JAX JIT is $6\times$ faster than DaCe AD, but this is only a difference between $35ms$ and $5ms$. For any array larger than that, DaCe AD is significantly faster, and the speedup difference only grows as the overhead from dynamic slicing larger arrays in JAX grows. For an array size of \textit{[1000x1000]} and $500$ timesteps, JAX JIT runs longer than 24 hours, the maximum time limit of an allocation on our computing nodes.

\subsection{GPU Performance Results}
\begin{figure}
    \centering
    \includegraphics[width=1\linewidth]{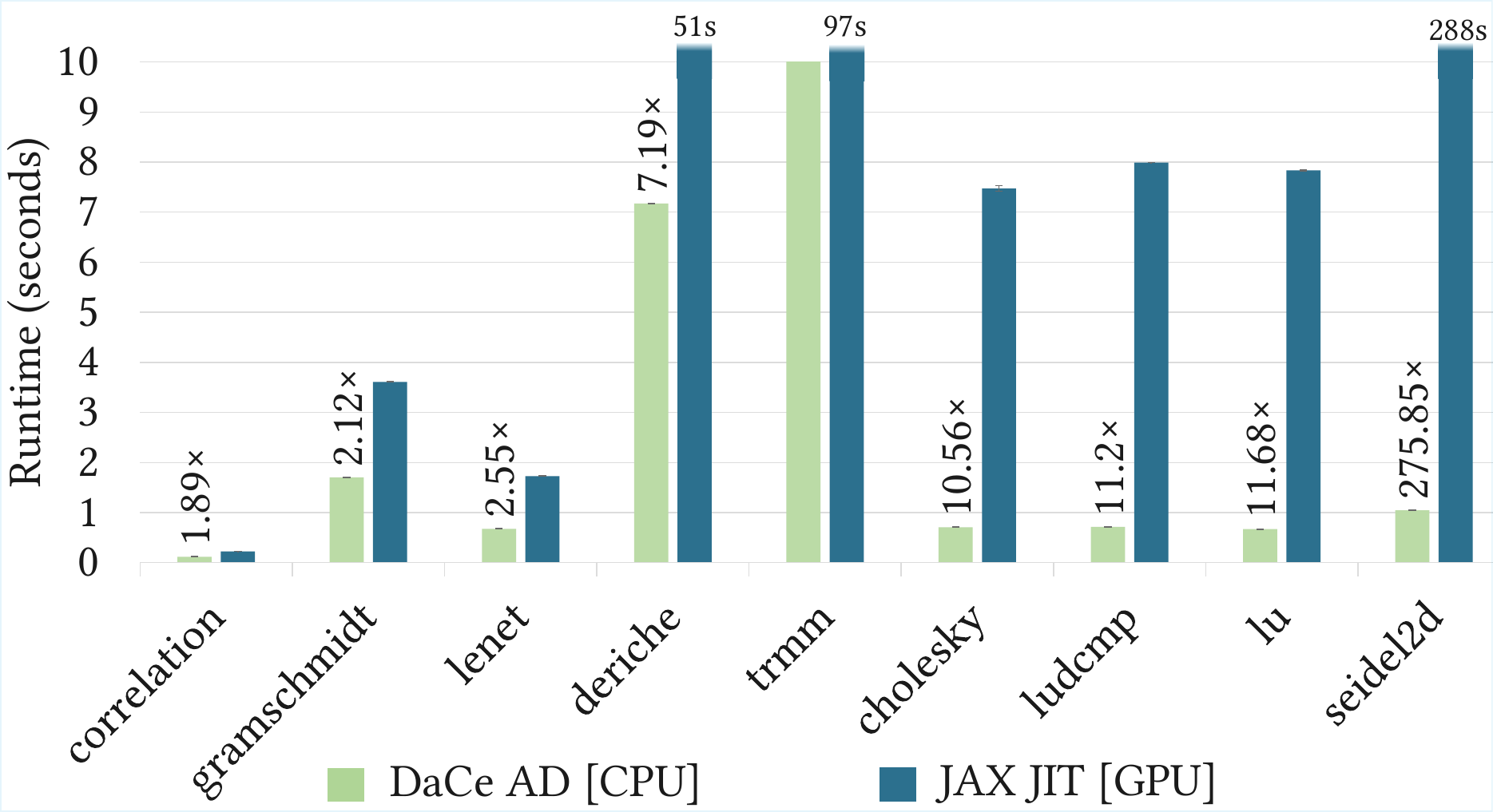}
    \caption{Performance results for DaCe AD [CPU: Intel Xeon Gold 6154] vs JAX JIT [GPU: NVIDIA V100]}
    \vspace{-0.4cm}\label{fig:gpu_results}
\end{figure}

The presented contributions of DaCe AD (efficient slicing and in-place gradient propagation) and JAX issues (dynamic slicing inefficiency, array immutability overhead, and extra bound checking) are algorithmic and persist for both CPU and GPU. We evaluate performance on an NVIDIA V100 for a subset of programs and find that JAX JIT improves performance in many cases. However, due to the large performance gap, this is not enough to beat even the DaCe CPU performance in 9 benchmarks. Figure \ref{fig:gpu_results} presents the results for these 9 cases. For example, on Seidel2d, the speedup of DaCe AD vs JAX goes from $2724\times$  to $275\times$ when switching from CPU to GPU. In other cases, running DaCe AD on the GPU yields even better performance compared to the CPU. For Jacobi2d, DaCe AD is slower ($0.85\times$) than JAX on CPU, while we are $1.26\times$ faster on GPU. Overall, the performance on GPU has a smaller gap, but DaCe AD still outperforms JAX on average.

\subsection{ILP Checkpointing}
\label{eval_ilp}

We evaluate the proposed ILP solution to the re-materialization problem.
Figure~\ref{fig:eval_ilp_example} compares the runtime and memory usage of all possible configurations for the program presented in Section~\ref{ilp_motivating_example}. As we need to decide on storing/recomputing three arrays, this translates to $2^3$ possible configurations. Our ILP-based solution automatically selects configuration C-3 from Figure~\ref{fig:eval_ilp_example}, which consists of storing $A_1$ and $A_2$ and only recomputing $A_0$. This is the fastest configuration that respects the memory constraint. Note that the peak memory measurements are adjusted by removing the program context overhead for all the measurements. Because the number of decision variables is very small (three), the time to solve the ILP is also negligible at $6.4$ms.

\section{Related Work}

Over the decades, many AD tools and approaches have been developed. The field of AD started with approaches such as ADIFOR~\cite{adifor}, a tool that differentiated Fortran 77 code in the 90's, and continued with similar approaches such as Tapenade~\cite{tapenade} and ADIJAC~\cite{adijac}. 


\subsection{Modern AD Tools for ML}    

AD tools experienced tremendous growth in popularity and adoption with the ML revolution, with many tools offering specialized, efficient support for gradient computation in this context.

\textit{PyTorch~\cite{pytorch}} is one of the most used frameworks for AI training. It is heavily optimized for ML applications for both performance and program support. Torch's engine has limitations when applied to scientific computing. In-place operations are not allowed for gradient-tracked arrays, and it is the user’s responsibility to decide what needs to be stored/recomputed. PyTorch supports an experimental AOT feature, which is a min-cut-based algorithm to minimize memory transfer in GPU kernels that relies on the fusibility of pointwise operations that are heavily present in DL models but are not centric to scientific computation. This method also relies on a hardcoded list of memory-bound operators to reduce memory usage, unlike our method, which automatically analyzes the memory patterns of general computations. 

\textit{TensorFlow~\cite{abadi2016tensorflow}} is a machine-learning library. Similar to PyTorch, TensorFlow uses a store-all strategy by default and has limited support for non-ML code patterns. 

\subsection{Modern General AD Tools}

Some tools offer support for a wider range of programs than simply ML codes, and are thus the closest comparison points to our work.     

\textit{Enzyme AD~\cite{enzymead}} is an automatic differentiation framework that works on the LLVM IR~\cite{LLVM} level. It can thus support a wide range of languages like C, C++, Swift, Julia, Rust, Fortran, etc. Enzyme is suitable for particular domains like climate modeling that rely heavily on Fortran code but limited for many domains that rely on the much more approachable Python from the non-computer-scientist domain specialists. By its nature, the tool cannot enable the interaction between scientific computing code and machine learning models, while DaCe AD provides a single environment where both code patterns can interact seamlessly. Because Enzyme AD is not an optimization framework, the gradient code's performance depends largely on the initial code's performance. This means that it is the user's responsibility to optimize performance. 

Later works on Enzyme AD ~\cite{enzymead2,enzymead3} added a store/recompute heuristic that first evaluated whether some values could be recomputed or not, then did a minimum-cut recompute vs. cache analysis that aims “to find the minimal set of values to cache by determining a minimum branch cut between values that must be cached and instructions that require values from the forward pass”~\cite{enzymead2}. This heuristic, however, does not set a specific memory limit under which to perform the optimization, unlike our solution.

\textit{Zygote~\cite{zygote}.}
Julia~\cite{julia} is a high-level, general-purpose programming language most commonly used for computational science. The Zygote~\cite{zygote} package implements reverse-mode AD for Julia programs. Zygote uses a store-all strategy by default. Other packages were implemented to support checkpointing strategies, most notably Checkpointing.jl~\cite{checkpointjl}, which supports several heuristics for recomputation. Scientists looking to use Julia for AD must rewrite their existing code in the programming language. Our solution aims to support existing scientific code, which is hard (and error-prone) to rewrite. Julia also lacks AD interoperability features with other languages, such as Python and Fortran.

\subsection{Prior Research on AD Checkpointing}

\paragraph{ML Focused solutions} Most research in this area has been focused on automating checkpointing in deep learning frameworks~\cite{suboptimal_nets, through_time,checkmate, how_to_train, monet}. These works, motivated by the pressing needs of the deep learning community, focus on reducing memory usage within the context of neural network training and inference. Among these, Checkmate~\cite{checkmate} is the most closely related to our approach: it leverages a mixed-integer linear program (MILP) to optimize memory usage in arbitrary deep learning graphs. Checkmate requires extensive operator-level profiling within TensorFlow to estimate recomputation costs. In contrast, our solution avoids profiling by computing costs through static analysis. Checkmate’s design assigns binary variables to each operator in a deep learning graph, causing the number of decision variables to grow rapidly for large models—sometimes leading to solver runtimes of several hours~\cite{checkmate}. Our design assigns decision variables at the level of array containers instead, limiting the variable count and reducing solver overhead. Finally, Checkmate is challenging to extend to the varied computational graphs found in scientific computing.

\textit{Revolve~\cite{revolve}} is an influential early work on AD checkpointing that provides an optimal checkpointing strategy for sequential or time-stepping computations. Its main strength lies in the theoretical guarantees it offers for minimizing recomputation when limited checkpoint storage is available. However, Revolve assumes a single sequential loop, uniform checkpoint costs, and known iteration counts, and thus cannot handle more complex computation graphs containing branching or varying costs. Moreover, its implementation can require intricate manual modifications that are challenging to maintain in large code bases. Our approach supports branching and uses dataflow analysis to determine costs automatically, making it suitable for more general computational structures.

\section{Conclusion}
In this work, we present DaCe AD, an automatic differentiation tool supporting both machine learning and scientific computing programs. DaCe AD employs a novel, efficient gradient calculation for loops that outperforms JAX JIT, a state-of-the-art AD framework, by up to three orders of magnitude at best, and by a factor of 4.1 across the  NPBench suite as a whole. We further introduce a novel automatic checkpointing strategy that automates the store/recompute decision given a user-defined memory limit, thus empowering scientists to reap the benefits of AD in more complex programs than ever before, and explore new AI4Science hybrid AD applications.

\section*{Acknowledgment}
This work was supported by the ETH Future Computing Laboratory (EFCL), financed through a donation from Huawei Technologies, and by the SwissTwins project (funded by the Swiss State Secretariat for Education, Research and Innovation) and the ERC PSAP project (Grant Agreement No. 101002047). We also acknowledge the Swiss National Supercomputing Centre (CSCS) for providing computational resources and technical support that facilitated this project.

\bibliographystyle{IEEEtran}
\bibliography{references}

\end{document}